\documentclass[conference]{IEEEtran}

\usepackage{cite}
\usepackage{amsmath,amssymb,amsfonts}
\usepackage{algorithmic}
\usepackage{graphicx}
\usepackage{textcomp}
\usepackage{textgreek}
\usepackage{xcolor}
\usepackage{url}
\usepackage{siunitx}
\usepackage{newunicodechar}
\usepackage{multirow}
\usepackage{booktabs}
\usepackage{threeparttable}
\usepackage{makecell}
\usepackage{array}
\usepackage{caption}
\usepackage{adjustbox}
\usepackage{pgf}
\usepackage{colortbl}

\newcommand{\heat}[1]{%
  \cellcolor{blue!\number\numexpr 100*#1/28\relax}%
  \ifnum\numexpr 100*#1/28\relax>55\color{white}\fi
  #1%
}

\newunicodechar{≈}{{$\approx$}}
\def\BibTeX{{\rm B\kern-.05em{\sc i\kern-.025em b}\kern-.08em 
    T\kern-.1667em\lower.7ex\hbox{E}\kern-.125emX}}

\title{\textit{NeuroHex}: A Brain-Inspired Hex Coordinate System to Enable Highly Computationally-Efficient World Models for Continuous Online-Adaptive  Learning}
\author{
\IEEEauthorblockN{Quinn Jacobson, Joe Luo, Jingfei Xu, Shanmuga Venkatachalam, Kevin Wang, Dingchao Rong, John Paul Shen}
\IEEEauthorblockA{\textit{NeuroAI Computer Architecture Lab (NCAL), Carnegie Mellon University}\\
  \{\textit{qjacobso, jpshen}\}\textit{@andrew.cmu.edu}}
}

\begin{document}
\maketitle

\begin{abstract}
\textit{NeuroHex}\footnote{This is an expanded version of the paper titled \textit{"NeuroHex: Highly-Efficient Hex Coordinate System for Creating World Models to Enable Adaptive AI"} published in the proceedings of the 2026 Neuro Inspired Computational Elements (NICE)\cite{jacobson2026NeuroHex} conference. This is an archival version of the paper and is currently under review for an ACM journal publication.} is a brain-inspired hexagonal coordinate system designed to support highly efficient world models and reference frames for online adaptive AI systems. 
Inspired by the hexadirectional firing structure of grid cells in the human brain, NeuroHex adopts a cubic isometric hexagonal coordinate formulation that provides full 60° rotational symmetry and low-cost translation, rotation and distance computation. We develop a mathematical framework that incorporates ring indexing, quantized angular encoding, and a hierarchical library of foundational, simple, and complex geometric shape primitives. These constructs allow low-overhead point-in-shape tests and spatial matching operations that are expensive in Cartesian coordinate systems. To support realistic settings, we also develop a novel tool (\textit{OSM2Hex}) that can process OpenStreetMap (OSM) data sets and convert them into the NeuroHex coordinate system. The OSM2Hex spatial abstraction processing pipeline can achieve a reduction of 90-99\% in geometric complexity while maintaining the relevant spatial structure map for navigation. Our initial results, based on actual city and neighborhood scale data sets, demonstrate that NeuroHex offers a highly efficient substrate for building dynamic world models to enable adaptive spatial reasoning in autonomous energy-efficient AI systems with continuous online-adaptive learning (COAL) capability.
\end{abstract}


\section{Introduction}
The biological brain has served as an inspiration for artificial intelligence (AI) and a proof-point for what can be theoretically achievable. AI and generative AI have made great progress in the last decade.
Although current AI approaches are impressive, they fall short in many ways in comparison to what biological brains can achieve. There are numerous efforts to create better AI, with one approach focusing on leveraging neuroscience insights to create fundamentally new AI approaches that incorporate world models and reference frames for online adaptive AI systems \cite{taniguchi2023world}. 

The heart of our approach is to reverse engineer how the biological brain learns and organizes information to enable efficient spatial recognition and reasoning, and at the same time online adaptive learning capability. Some key hypotheses  underlie our approach. Our first hypothesis is that the foundational building blocks of our long-term memory consist of many small reference frames where information is organized according to spatial relationships.
This is inspired by Hawkins and the Thousand Brains Project \cite{hawkins2017theory} \cite{hawkins2021thousand}. These reference frames must be optimized for creating hierarchical organizations of information and must be extremely computationally efficient. 

Our second hypothesis is that our working memory is based on a world model that handles hierarchical information and synthesizes inputs from multiple sensory modalities \cite{quak2015multisensory}. Our sensory input and physical interaction with the world are based on an egocentric coordinate system, while our world model and reference frames are based on allocentric coordinates. It is important that any spatial encoding of information supports efficiently converting between egocentric (polar coordinates) and allocentric (coordinates relative to local landmarks). It is also important that we can efficiently match observed objects with objects stored in our short-term and long-term memories When matching objects, the orientation and size we observe may differ from what we have in our memory.

In this work, we limit the scope of our work to learning, organizing, and recognizing spatial information in the physical world. We want to demonstrate this before extending to more general learning and handling of abstract concepts. We also focus initially on two dimensional information. For building mental maps of the physical world and our understanding of where we are in the physical world, we emphasize position in a horizontal plane. 
Based on how biology has solved these same problems and our observed computational costs, working in a hexagonal coordinate system provides many benefits. We start with background information and provide the evidence that biology actually uses a hexagonal coordinate system \cite{hafting2005microstructure, sargolini2006conjunctive}. Next, we cover 
our simple model of the brain and the key operations we are implementing as motivation for adopting a hexagonal coordinate system. We then discuss the specific hexagonal coordinate system and spatial building blocks that we have developed in \textit{NeuroHex}.
Finally, we explore how real spatial data on city scales can be encoded in NeuroHex and develop a conversion tool we called \textit{OSM2Hex}.

\section{Background}

Recently, leading experts in both communities have suggested the need to (re)connect research in Neuroscience and Artificial Intelligence \cite{zador2023catalyzing} to accelerate the development of next-generation AI systems with more brain-like capabilities and efficiency. They term this two-way convergence: \textit{NeuroAI}. We fully embrace this new vision and extend it to also include hardware fabric in a three-way convergence of Neuroscience, AI, and Computer Systems with the goal to build next-generation \textit{"NeuroAI Computing 
Systems."}

\subsection{Cortical Columns and Reference Frames}

Our NeuroAI approach is influenced by the work of Jeff Hawkins and Numenta \cite{hawkins2017theory}, which was based on the discovery by Vernon Mountcastle of the columnar organization of the neocortex \cite{mountcastle1997columnar} and the theoretical neuroscience work on reference frames \cite{barsalou1999perceptual} \cite{petersen2015representation}. Hawkins' Thousand Brain Theory \cite{hawkins2021thousand} suggests that \textit{Cortical Columns} (CC) are the fundamental neocortical units that embody intelligence. CCs function as \textit{Reference Frames} (RFs) that provide a sensorimotor modeling system. RFs record information through movement (sensorimotor learning), as the associated sensing agent moves through an environment, which could be physical or conceptual.

In Hawkins' model, multiple RFs targeting different sensory modalities can interact with each other and seek consensus on the output through voting within and between sensory modalities. In contrast to current DNNs that separate training and inference, CCs that store and process information in RFs can support real-time inference, as well as online, concurrent and continuous learning, and dynamically adapt to changes in sensory input from the environment.
The continuous feedback loop of predict-sense-update within a reference frame effectively introduces an additional dimension of "memory" that remembers past observations and patterns. This "sequential" (or recurrent) behavior is missing in all feed-forward neural networks, e.g. CNNs and DNNs. We believe  using reference frames with their associated world models, we can achieve continuous online-adaptive learning (COAL) capability.

\subsection{Hexagonal Coordinate Systems} \label{HexCoordinateSystems}

\subsubsection{Neuroscience Background and Inspiration}
The concept of navigation in mammals encompasses a wide range of sensory processing. The brain consolidates the perception of self-location, distance and direction with a topographically organized mapping of spatial data within grid cells in the medial entorhinal cortex (MEC) \cite{hafting2005microstructure}. Structurally, the entorhinal cortex is located at the interface between the neocortex and the hippocampus. MEC contributes to the formation and reinforcement of allocentric spatial memories, while the lateral EC (entorhinal cortext) carries context information and enables processing of temporal events \cite{mcnaughton2006path,solstad2008representation,hoydal2019object}. This division of labor supports activities such as path-finding, landmark anchoring, and reference frame translation. 

\textit{Grid Cells} function as a type of spatial correlation that establishes the sense of distance and direction over time. They are discharged at multiple locations repetitively to symbolically explore the mental environmental map, reinforcing and memorizing the image. They also receive feedback from the hippocampus through the subiculum and update themselves routinely to integrate positional and directional information \cite{taube2007head}. Deeper inspection of firing rates frequently yields a two-dimensional spatial autocorrelation function that exhibits distinct six-fold (\ang{60}) symmetric peaks \cite{jacobs2013direct}. 

Grid cells operate like an embedded coordinate system that provides a repeating lattice of activities to allow animals to estimate their relative positions with external cues. The discrete grid cell modules align with the environmental geometry by spontaneous and individualized updates corresponding to positional, orientational, and boundary information \cite{stensola2012entorhinal} \cite{krupic2015grid} and interact with hippocampal \textit{Place Cells} during remapping \cite{fyhn2007hippocampal}. Together, they form a flexible and stable  frame of reference that adapts to new environments while preserving previously learned maps. In humans, evidence from functional magnetic resonance imaging(fMRI) also indicates a hexadirectional (six-fold rotational symmetric) firing patterns of grid cells in EC during the navigation of a virtual environment \cite{doeller2010evidence} \cite{maidenbaum2018grid}. In addition to spatial cognition, a hexagonal grid-shaped signal is also capable of coding abstract knowledge and experiences \cite{constantinescu2016organizing}.

\begin{figure*}
    \centering
    \includegraphics[width=\linewidth]{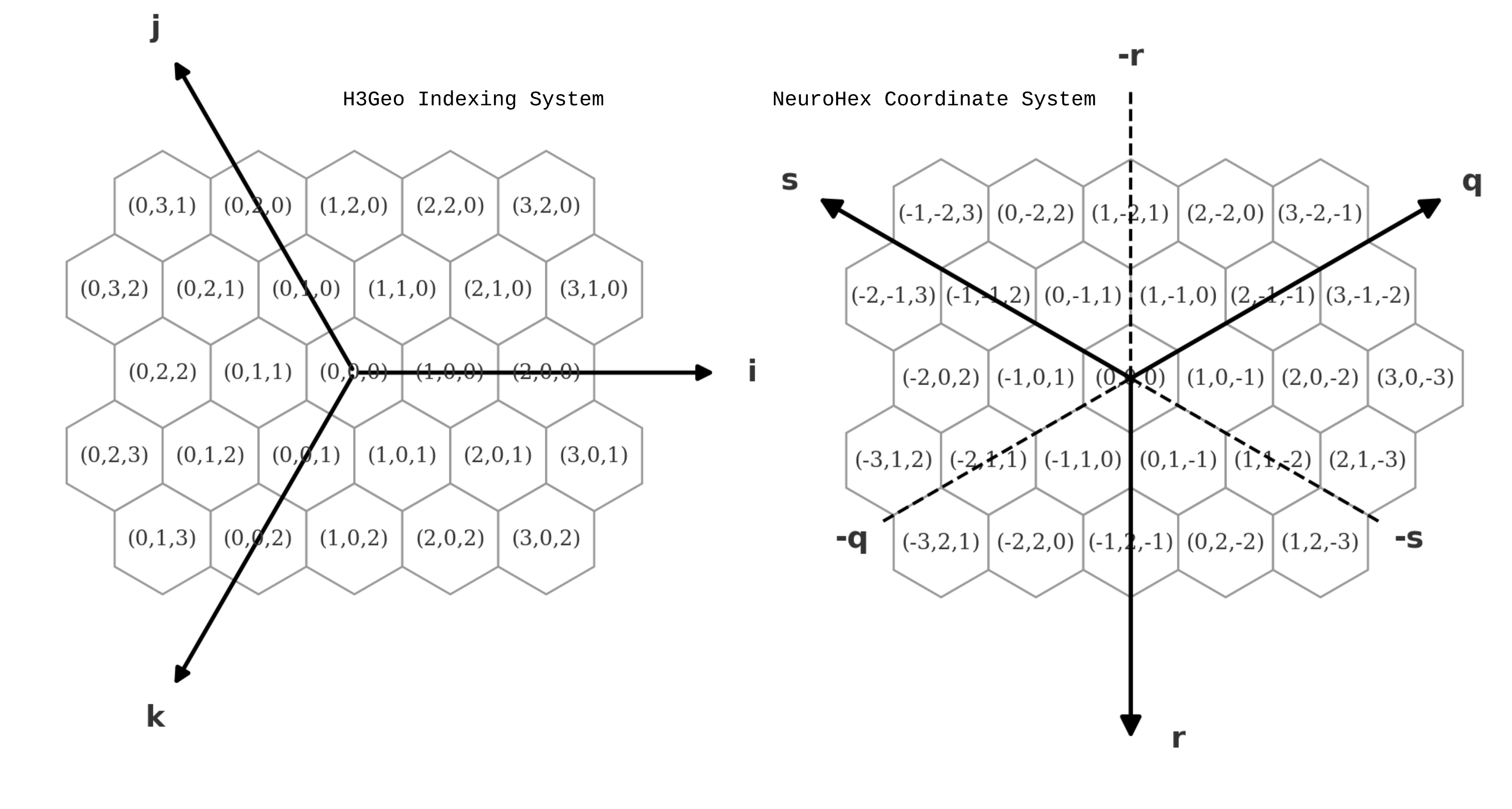}
    \caption{\centering Hexagonal Coordinate Systems: H3Geo (left) with positive coordinates (i,j,k) \cite{H3Geo}; our NeuroHex (right) with positive and negative coordinates (q,r,s).}
    \label{fig:H3vsNeuroHex}
\end{figure*}

\subsubsection{Current Hexagonal Coordinate Systems}
Hexagonal spatial data require a coordinate system to numerically represent locations within the grid. The H3Geo framework \cite{H3Geo} (see Figure \ref{fig:H3vsNeuroHex}), which represents the Earth’s surface using a hierarchy of mainly hexagonal cells projected onto the faces of an icosahedron, employs a $(i,j,k)$ coordinate system defined along three 120° axes, normalized so that all coordinates are positive and one coordinate is always zero. 
With the $(i,j,k)$ coordinate system, translation by a vector requires normalization to preserve these constraints, and the lattice exhibits symmetry only every 120°. Each cell is assigned a unique integer index that encodes its hierarchical resolution and location within each parent cell, with lookup tables used to convert between index values and the base coordinates $(i,j,k)$. Although this indexing system supports the transition between resolutions, it introduces additional computational overhead for normalization and index-to-coordinate conversion.

An alternative is the isometric cubic coordinate system, which is what NeuroHex is based on. (see Fig. \ref{fig:H3vsNeuroHex}) It is a fully symmetric representation of the hexagonal grid \cite{her1992symmetrical}. In this model, each hex cell is assigned a triplet $(q,r,s)$ that resembles cubic coordinates on the plane $x + y + z = 0$ in a 3D Cartesian space. This formulation preserves all six directions of movement and achieves full 60° of symmetry. Such cubic coordinates require no normalization after translation operations, since addition and subtraction of coordinate triplets automatically preserve the zero-sum invariant. The redundant third axis for our 2D hexagonal grid enables highly efficient algorithms to calculate movement, rotation, and distance.

\section{Working Model of The Brain and Motivation for NeuroHex}

This section introduces a high-level working model of the brain to provide context and motivation for NeuroHex. Our working model is based on the interpretation of neuroscience through the lens of trial and error simulation. We used a mouse in a maze (both constrained with walls and open fields) to develop an architecture where the mouse can learn, recognize, and navigate many mazes. We know that our brain model and reference frame (RF) organization are not exactly equivalent to biology, but we found them to be a useful abstraction.

\begin{figure*}
    \centering
    \includegraphics[width=\linewidth]{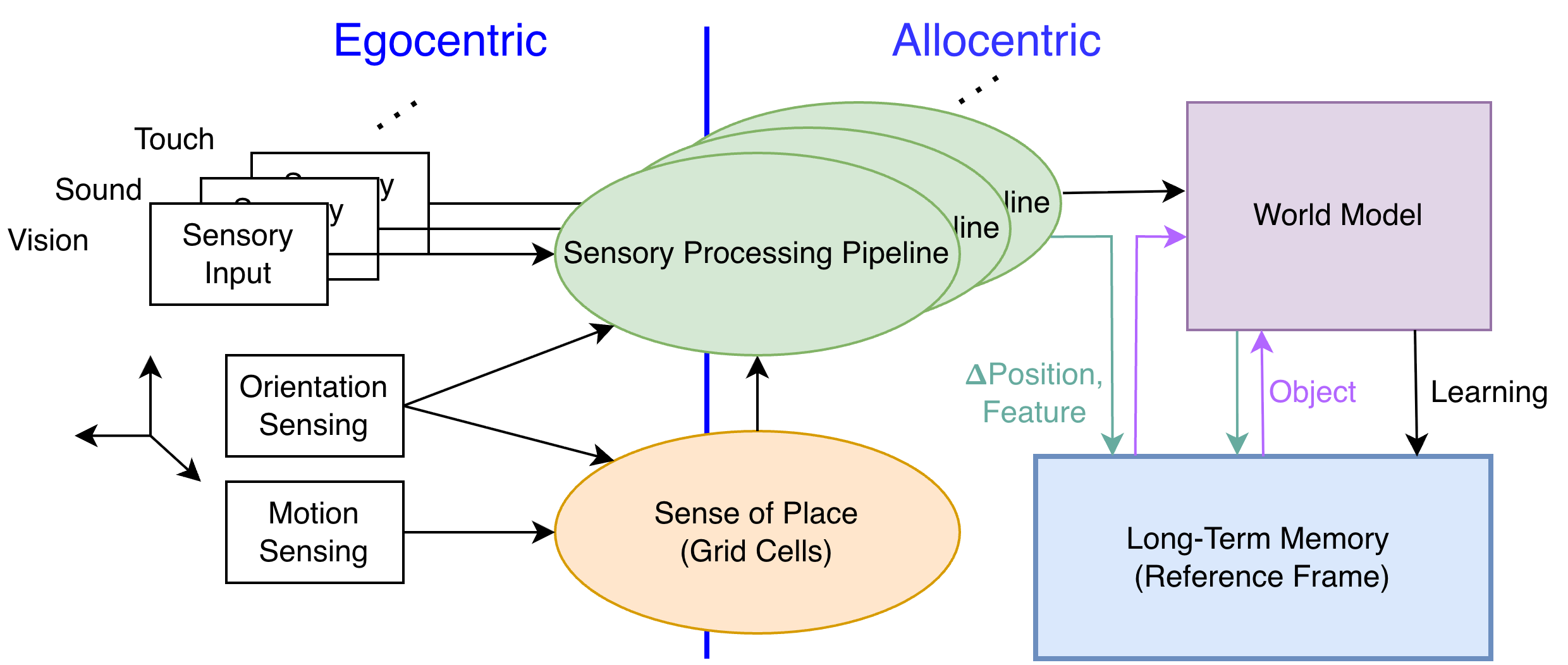}
    \caption{\centering High-Level working model of the brain with both Egocentric and Allocentric perspectives involving Grid Cells, adaptive World Models, and hierarchy of Reference Frames.}
    \label{fig:Top-Level Model}
\end{figure*}

Neuroscience research suggests that mammals have the tendency to form cognitive maps of what is present in the environment and where the elements of the environment are located in relation to each other. \cite{tolman1948cognitive}, \cite{o1978hippocampus}. 
Figure \ref{fig:Top-Level Model} shows a high-level brain working model. The ability to sense our orientation and movement in space helps us maintain a sense of place and is encoded in the grid cells. Sensory input, like vision, comes into the brain from an egocentric perspective. The sensory input needs to be processed into higher level features and moved into an allocentric coordinate system. Without moving into an allocentric coordinate system we would be overwhelmed as we moved through an environment, and everything constantly shifts in the egocentric perspective. Features and their relative positions are used to reference long-term memory and recognize known objects that can be put into a world model working memory. Objects in the world model and their relative positions can also be used to reference long-term memory to recognize higher-level objects. For learning, new objects can be constructed in the working memory and can ultimately be used to create new long-term memories.

\begin{figure}
    \centering
    \includegraphics[width=1.035\linewidth]{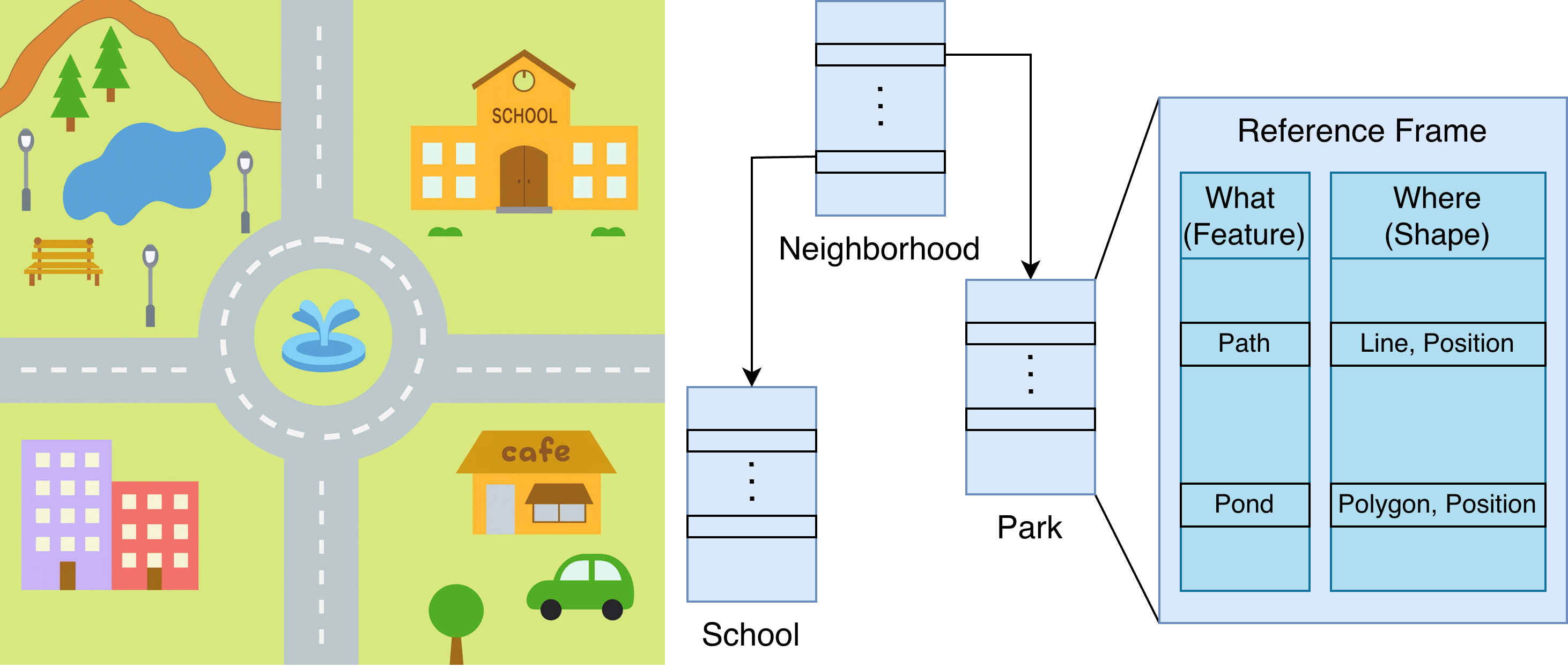}
    \caption{\centering Hierarchical Reference Frames used for building maps of physical world based on diverse sensory modalities.}
    \label{fig:HRF maps for phy. world}
\end{figure}

Long-term memory is a collection of reference frames. The heart of a reference frame is an associative structure that maintains a map of features and their relative positions, see Figure \ref{fig:HRF maps for phy. world}. The goal is to create a specific mental map of an environment, which is a hierarchical structure of reference frames, creating a 'what-where' representation of the physical world. Long-term memory is accessed by sending it a set of features with a relative (delta) position between them.

The reference frames in long-term memory attempt to match the set of features and their relative positions. To do this, each reference frame keeps track of a possible translation from the observed frame of reference to their local frame of reference. With multiple observations, based on possible translation matches, a reference frame gains confidence. The reference frames vote among themselves on the basis of their local confidence. An object is “recognized” when a reference frame reaches high confidence and responds with its local position translation (scaling, rotation, and position) and confidence.

We first consider the Cartesian coordinate system. If features are only points and we ignore rotation, the computational complexity is low. But to represent the world, we need a richer geometry including: straight lines, bent lines, polygons, and circles. As we extend the richness of the geometry, the computational cost of the matching operations becomes very expensive. Also, allowing objects to be matched in different rotational orientations makes the computational cost using traditional Cartesian coordinates even higher. 
While current GPUs and TPUs are quite effective for such computations, the question is whether we can achieve much better energy efficiency by adopting and leveraging insights from biology, including the use of hexagonal coordinate system.  

We strongly believe in the need to find a more efficient way to support spatial matching operations in world models of reference frames. We also want to find a solution that would support the efficient transition between egocentric (polar coordinates) and allocentric coordinates, which is essential to support continuous online adaptive learning. Based on inspiration from neuroscience, the isometric hexagonal coordinate system emerges as the ideal candidate.

\section{NeuroHex: Computationally Efficient Hexagonal Coordinates Systems}

Neither the Cartesian coordinate system nor the polar coordinate system offers the flexibility and computational efficiency required by our reference frames' architecture and implementation. We seek a coordinate system that incorporates both allocentricity and egocentricity in its location mapping and can efficiently implement proximity look-up, locational context switching, and path-finding with computational ease.

\textit{NeuroHex} is a novel hexagonal coordinate system for world models. It employs the world-centered model, with all directions having equal neighboring structures, to encode space isotropically, enabling the origin to be shifted with minimal effort. The spatial frame of hexagonal coordinates retains the relativity of individual hex grid cells by allowing for translation of magnitude and direction through simple vector addition and angle quantization. The hexagonal lattice structure preserves the allocentric attributes of Cartesian coordinates while simultaneously leveraging the computational efficiency associated with polar coordinates. 

\subsection{Mathematical Foundations of the Hexagonal Space}
To convert the firing patterns of grid cells into a tangible mathematical model for generalized usage, we represent the hexagonal pattern with tightly knitted flat-top or pointy-top oriented hexagons. An axial (cubic) coordinate system is used to enable standard mathematical operations on the hex grid \cite{patel2013hexagonal}, where all hex cells are denoted using cubic coordinates \((q,r,s)\), and are subject to the constraint: $q + r + s = 0$. NeuroHex adopts a flat-top hexagonal canvas, which can be partitioned into six \emph{wedges} made up of equilateral triangles that constitute a regular hexagon in clockwise order. \emph{Wedges} establishes the rudimentary principles of clockwise and counterclockwise angle reference for NeuroHex (Figure \ref{fig:NeuroHex Intro}). 

\subsection{NeuroHex Ring Encoding, Positional, Angular Arithmetics, and Sign-based Wedge Indexing}
The hexagonal canvas can be conceptualized as a central origin surrounded by successive concentric hexagonal rings of adjacent coordinates radiating outward. To effectively calculate displacement and angle within the hex space, we formulate a \emph{Ring Encoding} strategy for NeuroHex to represent the relative position of hex coordinates.

\begin{figure}
    \centering
    \includegraphics[width=0.95\linewidth]{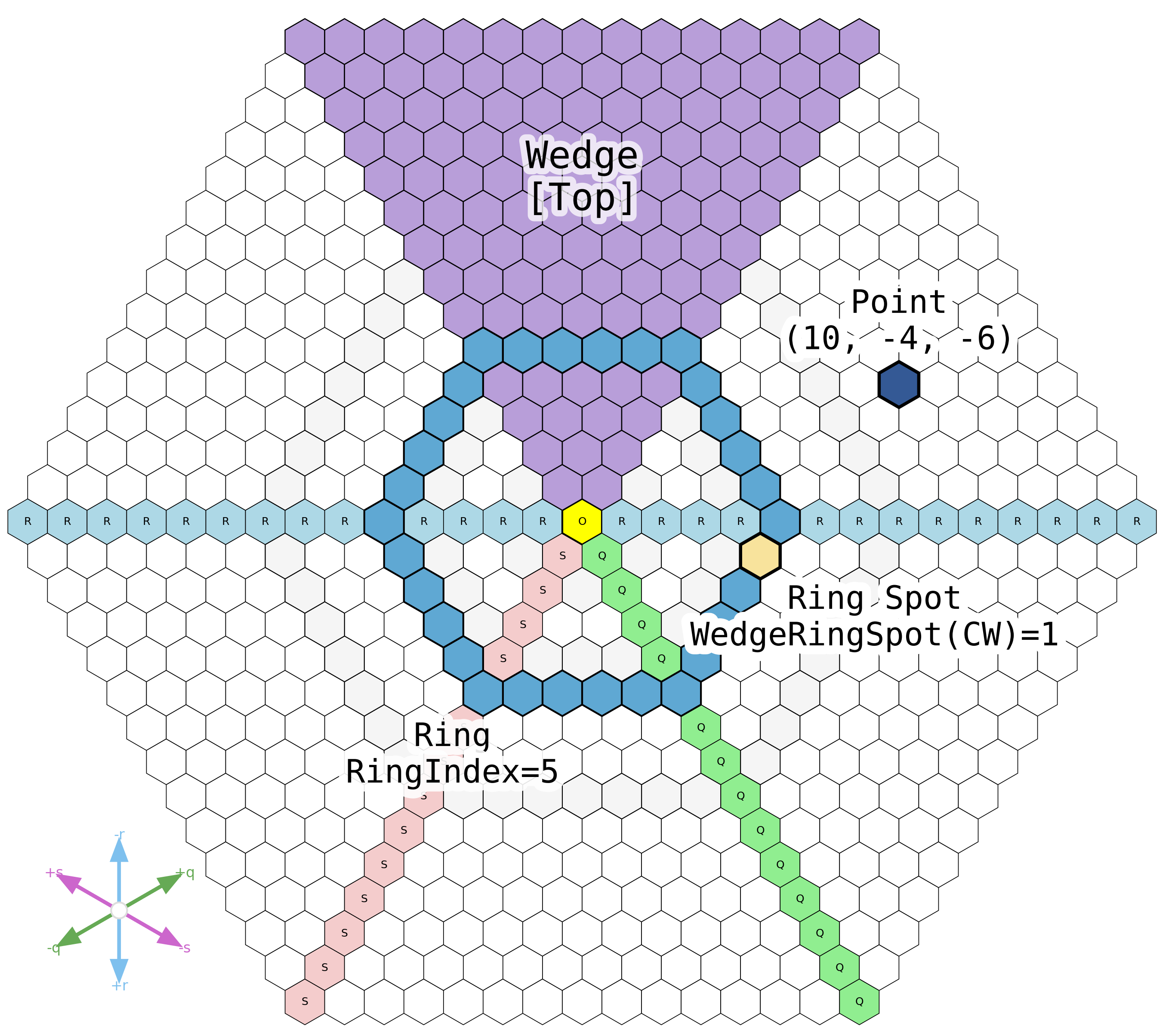}
    \caption{\centering Foundational Components in the NeuroHex coordinate system: Ordered Wedge, Ring, and Point.}
    \label{fig:NeuroHex Intro}
\end{figure}

\emph{Ring Encoding} assigns a $Ring Index(RI)$ to every ring to identify itself relative to an arbitrary origin, and a $Ring Spot$, which indicates the position of the coordinate on the designated ring. We define that the origin always starts at $RI$ "\textbf{0}". A full ring can be subdivided into 6 segments with inclusive endpoints that correspond to the 6 wedges of a hexagon. Within the ring segment of a wedge, any $Ring Spot$ will function as a relative $WedgeRingSpot$, effectively limiting its range from 0 to $RI$ if both axis lines are included in a wedge sector. An illustration of the basic components of NeuroHex is shown in Figure \ref{fig:NeuroHex Intro}. 

\begin{figure}
    \centering
    \includegraphics[width=0.8\linewidth]{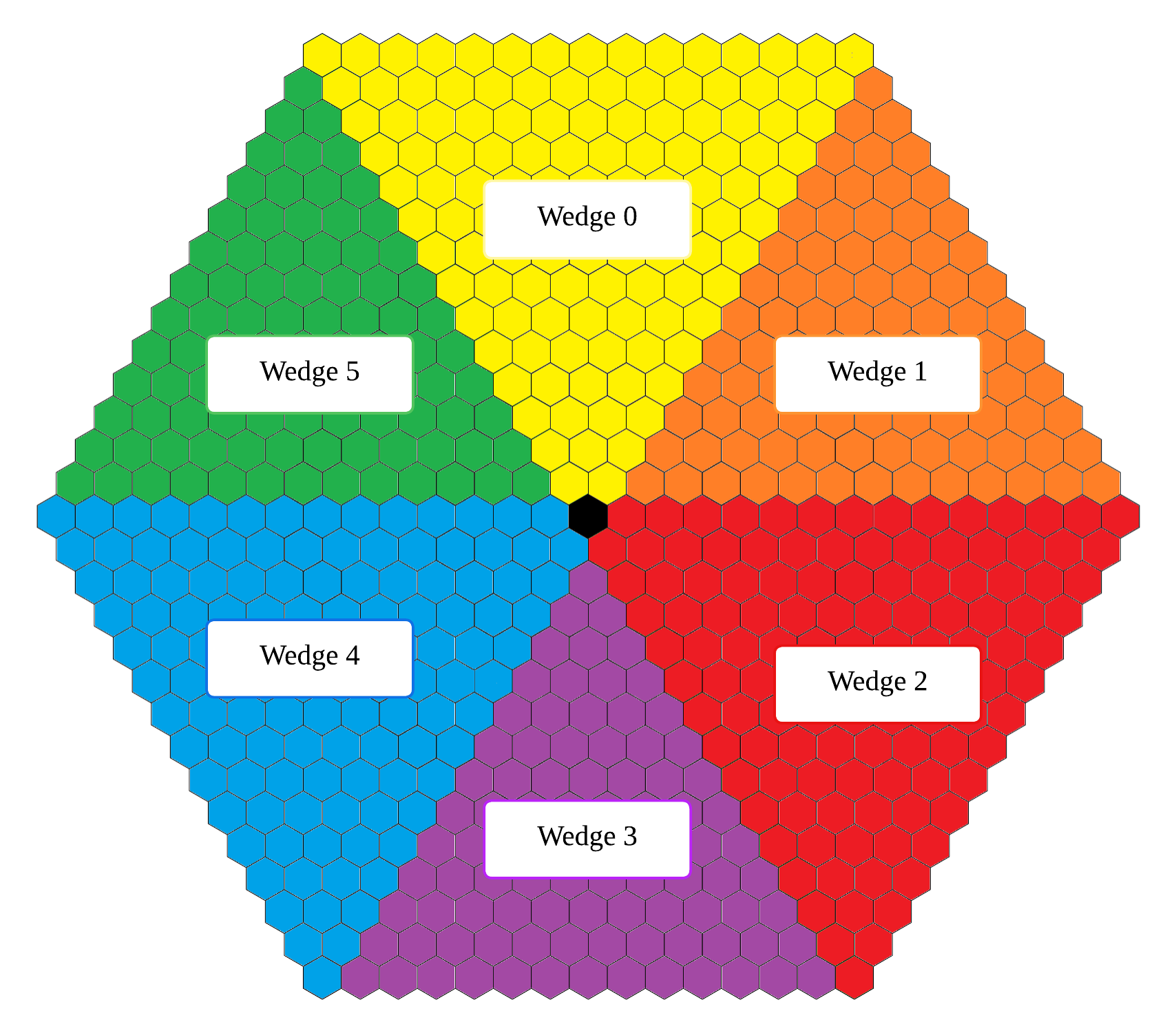}
    \caption{\centering Wedge Partitioning Based On Sign-based Wedge Indexing}
    \label{fig:SBWedgeIdx}
\end{figure}

\begin{figure}
    \centering
    \includegraphics[width=1.045\linewidth]{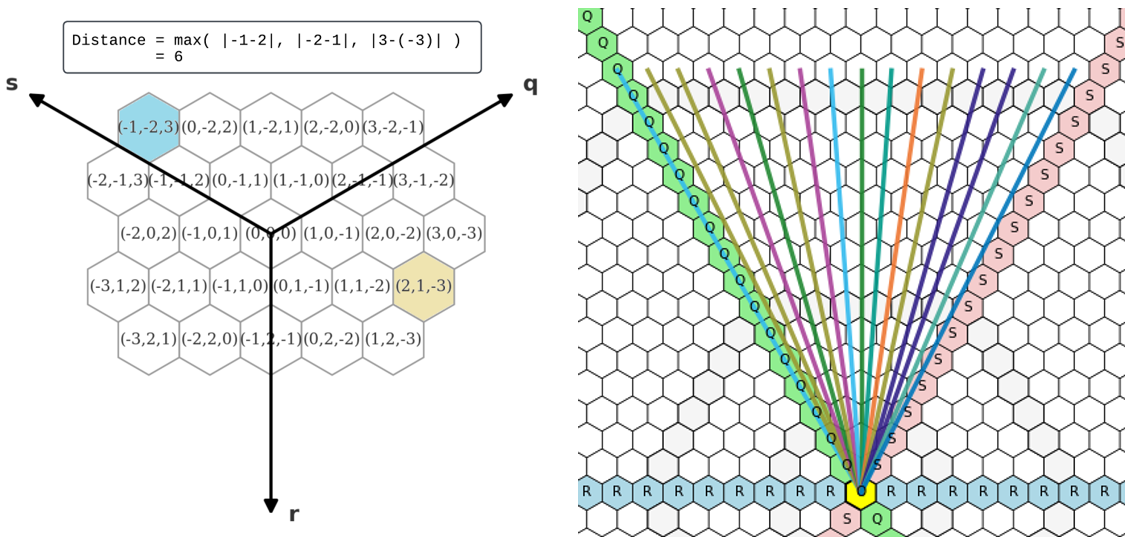}
    \caption{\centering Distance Computation (left) and Angular Direction Quantization (right) in the NeuroHex coordinate system.}
    \label{fig:Distance_and_Angle}
\end{figure}

NeuroHex coordinate system, derived from hexadecimal coordinates, possesses some other distinctive structural properties that make it particularly well-suited for geometric computation. The distance between two coordinates can conveniently be computed by taking the maximum of the absolute values of the differences of each cubic coordinate. Any coordinate on a ring is also $RI$ away from the origin. See Figure \ref{fig:Distance_and_Angle} (left). Defined by an origin and six relative wedges, the hexagonal canvas provides a seamless and efficient transition between egocentric and allocentric representations of the 2D world. Each hex coordinate consists of three signed values that uniquely identify the wedge in which they reside. There exist two indexing methods for NeuroHex wedges: \textbf{order-based} and \textbf{sign-based} indexing. Order-based indexing method labels wedge from 0 to 5 following clockwise order, with the top wedge being index 0 as shown in Figure \ref{fig:NeuroHex Intro}. The second indexing method characterizes them according to the non-negativity of their coordinate components. Starting from the top wedge counting clockwise, the six wedges can be identified by \(\{ \texttt{010}, \texttt{011}, \texttt{001}, \texttt{101}, \texttt{100}, \texttt{110} \}\) in binary by extracting the sign bits of the three coordinate numbers. The boundaries of the wedges are subsequently adjusted as shown in Figure~\ref{fig:SBWedgeIdx}. This encoding is called \emph{sign-based wedge indexing}. The redefinition of the wedge space and index eliminates the need to select the ordered wedge index through multiplexing, which is especially frequent within polar-related operations. The merged \textit{q}, \textit{r}, and \textit{s} axis line boundaries do not compromise the accuracy of wedge-based localization. For directional comparison, a coarse partitioning of the space is sufficient, while for operations outputting precise coordinate values, the sign-based wedge index is typically converted back into ordered wedge index, and the coordinates are usually denormalized prior to exact computation.

\emph{Directional computation} in NeuroHex can be achieved via angular quantization. 
Rotational references established by the positions of the wedges can be further quantized into arbitrary levels that define the resolution of the hex canvas and the geometric operations. Hexagonal rings formed by the coordinates surrounding the origin can be interpreted as anchor structures, with lines extending to the origin that partition each wedge into equal angular regions as seen in Figure \ref{fig:Distance_and_Angle} (right). For example, with this approach, assuming 4-bit quantization on every wedge, the maximum angular resolution for the entire canvas is $\frac{360}{(6\cdot16)} =$ \ang{3.75}. Both directional and angular transformations can be mapped to the reference ring to improve hardware performances, analogous to quantization of weights in machine-learning models. Within NeuroHex framework, such conversions are referred to as the normalization and denormalization processes. All geometric operations within the NeuroHex framework can be localized to a single wedge, and most reduce to the comparison of whether one coordinate on a given ring lies farther than another coordinate on a different ring. This constitutes the only operation in the NeuroHex coordinate system that requires multiplication; all other operations can be implemented using only addition and subtraction.

The computational benefits are not limited to the reduced multiplication counts. Divisions are excluded from all geometric operations, while the computation width is also effectively reduced. Assume an 8-bit wide signed NeuroHex canvas where possible coordinate values range from -128 to 127, localized geometric operations need only one additional bit for two’s complement addition and subtraction, and at most 18 bits for multiplication. Moreover, the inherent geometric structure of hexagonal grids avoids the use of square roots and trigonometric functions entirely. All wedge-related operations are minimal in complexity and require only 3-bit representations.

\subsection{NeuroHex Shape Library}
With the concepts of location and direction established, we can transform the hex coordinate system into a canvas decorated with "shapes" composed of hex coordinates labeled with various features, similar to rasterization. Drawing shapes in NeuroHex is equivalent to determining whether a given coordinate lies within a defined boundary. The process can be reduced to repeated assessments of a point's position relative to a directed line segment, known as \textit{orientation predicate} in computational geometry. 

\begin{figure}
    \centering
    \includegraphics[width=\linewidth]{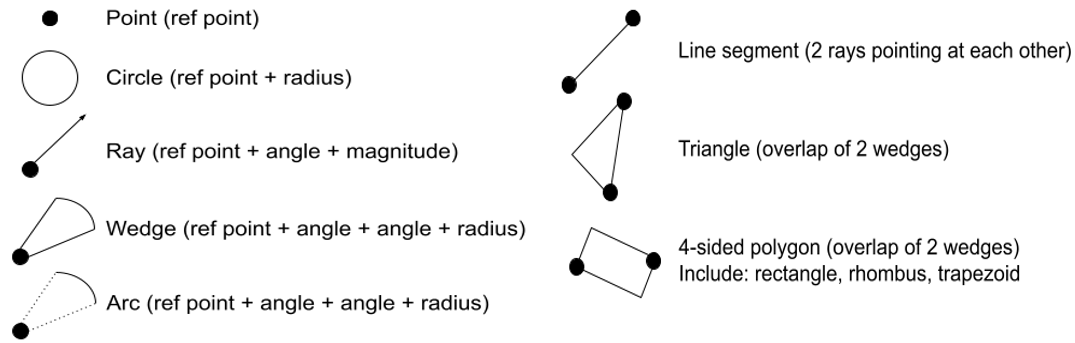}
    \caption{\centering Foundational Shapes (left) with single reference point, and Simple Shapes involving two reference points (right).}
    \label{fig:Basic_Shapes}
\end{figure}

We classify the NeuroHex shapes into three categories: \textit{Foundational}, \textit{Simple}, and \textit{Complex}, illustrated in Figure \ref{fig:Basic_Shapes}. 
\par
\subsubsection{Foundational Shapes}
A foundational shape is characterized by a hex coordinate, one magnitude, and two quantized angles.
The angles share a common vertex and are labeled to ensure that the latter angle is always situated in the clockwise direction relative to the former angle, allowing wrap-arounds at the edge of a full rotation. 

With a reference point, a magnitude, and two angles, it is possible to construct:
\begin{itemize}
  \item a \emph{point} by setting angles and magnitude to null.
  \item a \emph{ray} by setting one of the angles to null.
  \item a \emph{wedge} by setting all four properties to not null.
\end{itemize}
The catalog of foundational shapes is not restricted to these examples and can be customized as needed by altering and configuring what the null mask symbolizes.
\subsubsection{Simple Shapes}
A simple shape consists of the intersection of two foundational shapes. That is, a coordinate is considered to be enclosed in a simple shape if and only if it is bounded by both of the foundational shapes.
With such an intersection, we can obtain common shapes used for maps and computer graphics, such as:
\begin{itemize}
  \item a \emph{line segment} by rendering two rays facing each other.
  \item a \emph{triangle} by arranging two wedges with acute angles together such that they coincide along a common edge.
  \item a \emph{quadrilateral} like rectangle, rhombus, and trapezoid by positioning two wedges in opposing orientation.
\end{itemize}
A wide variety of other shapes can be built through the juxtaposition of foundational shapes. 
\subsubsection{Complex Shapes}
By integrating simple Boolean logic into the shape-generation process, we can construct various geometries through additive overlap$(AO)$ and subtractive overlap$(SO)$. These two operations serve as complements: $SO$ isolates coordinates contained by one shape but excluded from the other, whereas $AO$ selects coordinates jointly bounded by both shapes. Some of the shapes generated include donuts (subtractive of two circles) and tear drops (additive of two circles). Using more than two foundational shapes further expands the geometric shapes that can be represented.

Even with support for rich geometric shapes, the point-in-polygon test in NeuroHex requires only polar angle computations and logic operations and incurs substantially less complex computation than conventional computational geometry.

\subsection{NeuroHex Translation, Rotation, and Scaling of Shapes} 

\begin{figure}
    \centering
    \includegraphics[width=0.95\linewidth]{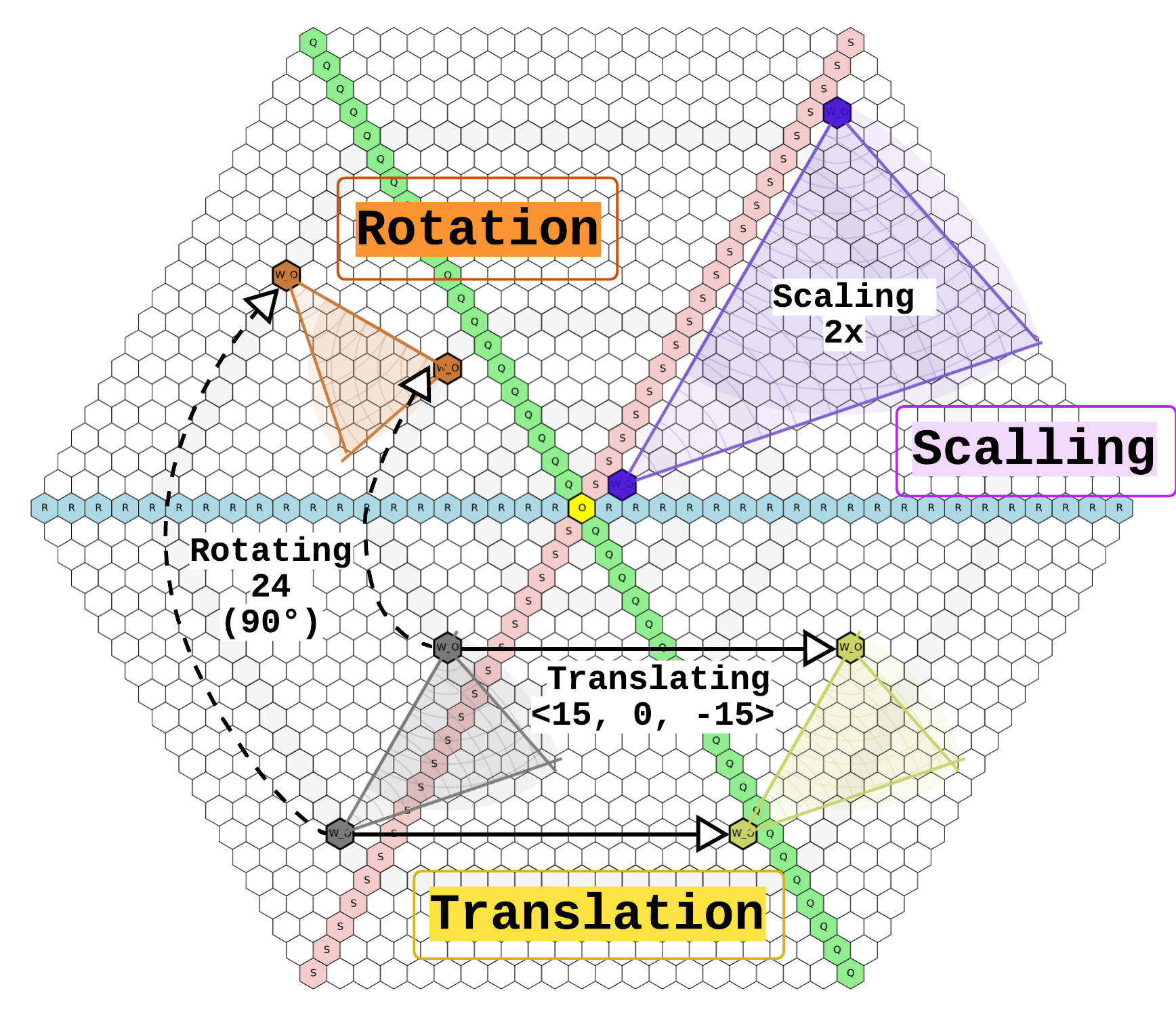}
    \caption{\centering Examples of Translation, Rotation, and Scaling of shapes in the NeuroHex coordinate system.}
    \label{fig:TRS_demo}
\end{figure}

Figure \ref{fig:TRS_demo} shows the efficient shape operations in NeuroHex.

\textit{Translation:} In addition to allowing for easy origin translation, NeuroHex also enables efficient translation of $magnitude$ and quantized angle input to cubic coordinates. $RingIndex$ is equal to $(magnitude+1)$. The quantized angle allows for straightforward computation of direction: let the maximum per-wedge quantization level be a power of 2, encoded using $n$ bits in binary, the higher order $(n+2)$ to $n$ bits of the angle encoding can be used to index the 6 wedges, while the lower $n$ bits represent the wedge-local angular quantization. Determining the exact $WedgeRingSpot$ requires a conversion from relative quantization to integer length, which involves a single instance of logical shift and multiplication. Once $RingIndex$, $WedgeIndex$, and $WedgeRingSpot$ have been derived, the target cubic coordinate resolution becomes trivial, since only the multiplexing selection of directions from the six wedges and potential sign reversal is needed without further arithmetic. This is illustrated in Figure \ref{fig:TRS_demo}.

\textit{Rotation:}  The rotation operation likewise benefits from the lightweight translation of polar inputs, since it merely appends the rotational offset to the quantized angle. As a result, the dominant computation cost of rotating a point around an arbitrary reference within the NeuroHex framework would involve 3-4 additions and a single multiplication. As previously described, NeuroHex builds simple and complex shapes from foundational shapes composed of point, angle, and magnitude. Rotating a shape around an arbitrary point is structurally equivalent to rotating the reference points and linearly translating the angles within the shape. An example of point rotation is shown in Figure \ref{fig:TRS_demo}. In comparison, an analogous operation in Euclidean space is considerably more complex, requiring rotation matrices with trigonometry.

\textit{Scaling:} Shape up-scaling in NeuroHex can be achieved through logical shifts of coordinates and magnitudes if the scaling factor is power of 2. Down-scaling requires an extra step of normalization (details covered in the next section) to maintain the constraints of the coordinate system. 

\subsection{Variable Resolution} 

\begin{figure}
    \centering
    \includegraphics[width=0.8\linewidth]{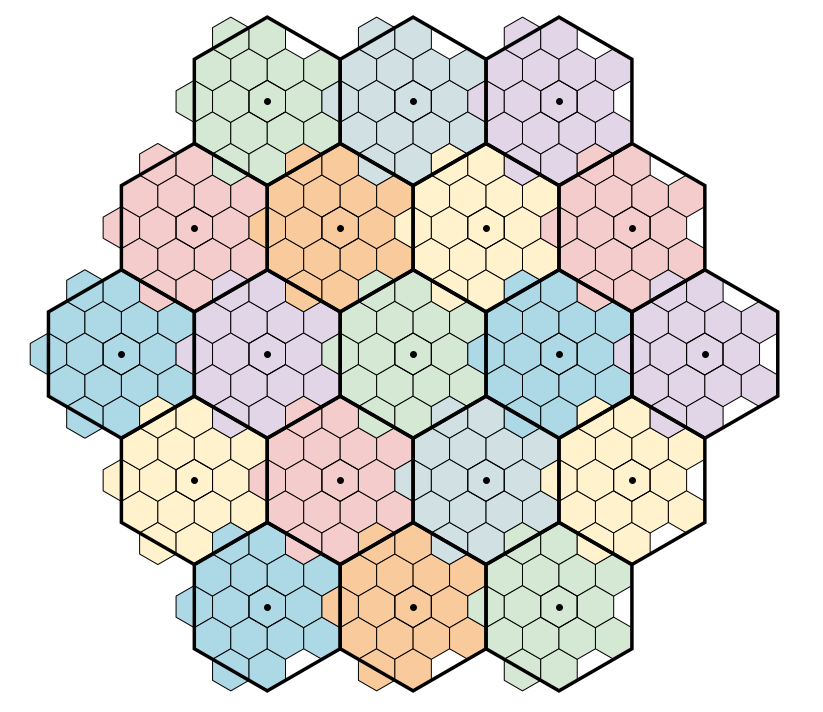}
    \caption{Variable Resolutions of Hexagonal Tiles in NeuroHex.}
    \label{fig:tiling}
\end{figure}

To gracefully handle working with objects of different sizes, we support different resolutions in NeuroHex. In a cubic-coordinate representation, multiplying the grid size by a factor $k$ is equivalent to tiling the plane with larger hexagonal cells whose centers lie at coordinates that are integer multiples of $k$. These centers maintain the exact geometric structure of the original lattice, which is important to reuse the existing arithmetic library. Limiting $k$ to a power of 2 reduces the implementation cost, as division and remainder extraction can be performed with bit shifts and bit masks in two’s-complement. Thus, to change to a coarser resolution, we must map a coordinate to the index of the coarser tile that contains it, as illustrated in Figure \ref{fig:tiling}. 

Dividing and truncating each coordinate by $k$ for $q$, $r$, and $s$ independently will lead to violations of the zero-sum constraint of the cubic coordinate system. After dividing each component by $k$, the sum of the 3 remainders can be represented as \( n \cdot k \), where \( n \in \{0,1,2\} \). To produce both valid cubic coordinate centers and tiles of equal size, we increment $n$ of the divided components that produce the $n$ largest remainder(s).

\subsection{Computational Efficiency of NeuroHex Coordinate System} 

Table \ref{tab:neurohex_comparison} compares the complexity of the computation for performing key geometric operations in the NeuroHex vs. Cartesian coordinate system and rasterization algorithm. The following is a major advantage: operations in the NeuroHex coordinate system avoid computationally intensive calculations. The required number of multiplications and divisions is limited and no trigonometric functions, that are often orders of magnitude more costly, are needed. For a world model defined within an N-bit resolution and range, NeuroHex requires at most $(N+B)$ bits to properly represent all features, since none exceed eight times the largest coordinate magnitude. In contrast, Cartesian coordinates inherently require up to $2N$ bits due to squaring. NeuroHex mitigates these complexities by employing much simpler ALU operations. Compared with rasterization rendering techniques commonly used in modern GPUs, NeuroHex eliminates the need for matrix arithmetic and edge functions, delivering superior efficiency with convenient and streamlined orientation determination. 
A much more in-depth analysis of computational complexity is presented in the next section. For each of the operations in Table \ref{tab:neurohex_comparison}, we quantitatively show the significant computational efficiency of spatial world models based on NeuroHex.

\begin{table*}[t]
\caption{Comparison of computation complexities between NeuroHex system and Cartesian computational geometry.}
\label{tab:neurohex_comparison}
\centering

\begin{threeparttable}
\footnotesize
\setlength{\tabcolsep}{3pt}
\renewcommand{\arraystretch}{1.02}

\begin{tabular}{|>{\raggedright\arraybackslash}p{0.18\textwidth}|
                >{\raggedright\arraybackslash}p{0.35\textwidth}|
                >{\raggedright\arraybackslash}p{0.4\textwidth}|}
\hline
\textbf{Operation} &
\textbf{NeuroHex Coordinate System} &
\textbf{Two-Dimensional Computational Geometry} \\
\hline

\makecell[l]{\textbf{\textit{Distance$^{*\textit{(a)}}$} \textbf{\textcircled{\raisebox{-0.9pt}{1}}}}} &
$\max\!\left(|q-q'|,\ |r-r'|,\ |s-s'|\right)$ &
$\sqrt{(x-x')^2+(y-y')^2}$ \\
\hline

\textbf{Radial Distance} &
$\max\!\left(|q|,\ |r|,\ |s|\right)$ &
$\sqrt{x^2+y^2}$ \\
\hline

\makecell[l]{\textbf{Denormalization}\\} 
&
$(R_{ws}\cdot R_i^{\textit{(b)}})\gg B^{\textit{(c)}}$ &
$\phi_{ref1} \cdot \phi_{ref2}\gg B$ \\
\hline

\makecell[l]{\textbf{Polar Angle $\mathbf{(\phi_{point}^{\textit{(d)}})}$ \textbf{\textcircled{\raisebox{-0.9pt}{2}}}}} &
$R_{ws}+W_i^{\textit{(e)}}R_i$ &
$\operatorname{atan2}(x,y)$ \\
\hline

\makecell[l]{\textbf{\textit{Sign Decoding$^*$ \textbf{\textcircled{\raisebox{-0.9pt}{3}}}}}} &
\makecell[l]{
$(q,r,s)=\bigl(s_q|q|,\ s_r|r|,\ s_s|s|\bigr)$\\
$\mathcal S(\text{Wedge})=(s_q,s_r,s_s)$\\
$s_q,s_r,s_s\in\{-1,0,+1\}$
} &
\makecell[l]{
$(x,y)=\bigl(s_x|x|,\ s_y|y|\bigr)$\\
$\mathcal S(\text{Quadrant})=(s_x,s_y)$\\
$s_x,s_y\in\{-1,0,+1\}$
} \\
\hline

\makecell[l]{\textbf{Orientation Predicate}\\} &
\makecell[l]{
Obtain $\phi_1$ and $\phi_2$ via \textcircled{\raisebox{-0.9pt}{2}}\\
and determine whether $\phi_1<\phi_2$
} &
$\det\!\begin{pmatrix}
x_1-x_2 & y_1-y_2\\
x-x_2   & y-y_2
\end{pmatrix}>0$ \\
\hline

\makecell[l]{\textbf{\textit{Point Translation$^*$}}} &
$(q-\Delta q,\ r-\Delta r,\ s-\Delta s)$ &
$(x-\Delta x,\ y-\Delta y)$ \\
\hline

\makecell[l]{\textbf{Point Rotation (CW)}} &
\makecell[l]{
1.\ Obtain $W_{p}$ and $R_{ws,p}$ via \textcircled{\raisebox{-0.9pt}{3}}\\[0.8ex]
2.\ Obtain $W_{rot}$ and quantized $R_{ws,rot}$ \\
3.\ $
W_{final}=
\begin{cases}
W_{p}+W_{rot}-6, & W_{p}+W_{rot}\ge 6\\
W_{p}+W_{rot}, & \text{otherwise}
\end{cases}$\\[0.8ex]
4.\ $
R_{ws, final} = R_{ws,p} + \text{denormalized} \space R_{ws,rot}.
$ \\ [1.1ex]
5. Trim overshot denormalized wedge ring from \\
$R_{ws, final}$. Increment $W_{final}$ if needed. \\ [0.8ex]
5.\ Assign position and sign for final coordinates.\\ 
} &
\makecell[l]{
$x'=x\cos\theta_{rot}-y\sin\theta_{rot}$\\[0.6ex]
$y'=x\sin\theta_{rot}+y\cos\theta_{rot}$
} \\
\hline

\makecell[l]{\textbf{Point-in-Sector Test} \textbf{\textcircled{\raisebox{-0.9pt}{4}}}} &
\makecell[l]{
Perform two \textcircled{\raisebox{-0.9pt}{2}} to obtain polar angles \\
of sector boundaries $\mathbf{\phi_{b1}, \mathbf{\phi_{b2}}}$ and denormalize.\\[0.6ex]
Perform one \textcircled{\raisebox{-0.9pt}{2}} to obtain polar angle $\mathbf{\phi_{p}}$\\
for the point and check if $\mathbf{\phi_{b1}} \le \mathbf{\phi_{p}} \le \mathbf{\phi_{b2}}$.
} &
\makecell[l]{
$\vec v=(x,y)$\\
$\vec a=(\cos\theta_1,\sin\theta_1)$\\
$\vec b=(\cos\theta_2,\sin\theta_2)$\\[0.4ex]
$\operatorname{InSector}(P)\iff
\left\{
\begin{array}{l}
\|\vec v\|^2\le R^2\\
\operatorname{cross}(\vec a,\vec v)\ge 0\\
\operatorname{cross}(\vec v,\vec b)\ge 0
\end{array}
\right.$
} \\
\hline

\makecell[l]{\textbf{Point-in-Triangle Test}} &
\makecell[l]{
Perform two \textcircled{\raisebox{-0.9pt}{4}} tests, since a triangle is\\
defined by two wedges.
} &
\makecell[l]{
Edge function \cite{pineda1988parallel}:\\
{\scriptsize
$E_{01}(P)=
(P_x-V_{0x})(V_{1y}-V_{0y})
-(P_y-V_{0y})(V_{1x}-V_{0x})$
}\\[0.4ex]
$\operatorname{InTriangle}(P)\iff
\left\{
\begin{array}{l}
E_{AB}(P)\ge 0\\
E_{BC}(P)\ge 0\\
E_{CA}(P)\ge 0
\end{array}
\right.$
} \\
\hline

\end{tabular}

\begin{tablenotes}[flushleft]
\footnotesize
\item[(a)] Marked operations are native to hexagonal grids. All other operations are developed in this work and are unique to NeuroHex.
\item[(b)] $R_i$ denotes the ring index. The wedge-local ring spot is $R_{ws}$.
\item[(c)] $B=\log_2(Q)$ is the bit-width of quantization levels, where $Q$ is the number of angular quantization levels per wedge.
\item[(d)] $\phi$ denotes the quantized clockwise angle in the NeuroHex representation.
\item[(e)] The six wedges of the flat-top hexagonal canvas are indexed by $W_i$ from 0 to 5, beginning at the top wedge and increasing clockwise.
\end{tablenotes}

\end{threeparttable}
\end{table*}

\section{Comparison of Hardware Implementations of NeuroHex vs. Cartesian Systems}

\subsection{Implementation Strategies for NeuroHex and Cartesian Geometric Operations}

\subsubsection{Distance}
Due to the nature of cube coordinates, distance calculation within NeuroHex can be simplified to the maximum of the absolute coordinate values, as shown previously. In the Cartesian coordinate system, the accurate distance must be calculated using a fixed-point approximation of the square root of the sum of squared differences.

\subsubsection{Polar angle}
The conventional way to determine relative angles formed by coordinate points in the Cartesian coordinate system is usually the arctangent function, but the full precision afforded by the trigonometric function is unnecessary. Instead, we implement a low-cost integer approximation of \texttt{atan2} that matches the angular granularity provided by the NeuroHex canvas. Given the input \textit{x} and \textit{y} coordinates, this method computes a quantized angle bin by quantizing the \ang{360} unit circle into \ang{4} snippets, thus avoiding trigonometry. 

With NeuroHex, polar angle mostly refers to the relative position of the point along a hexagonal ring. In subsequent operations where angle inputs are involved, most of the angular quantities are quantized with respect to rings whose scales are powers of two, which allows them to be mapped directly to the polar-angle representation through the denormalization procedure introduced earlier. The denormalization of degrees is a common recurring method employed throughout the implementations of other geometric operations mentioned in this paper.

\subsubsection{Rotation}
NeuroHex offers a major advantage in rotational operations by completely avoiding trigonometries. Through quantization, the angular resolution can be parameterized with respect to the rings scaled by powers of two. In this implementation, we adopt a 4-bit quantization scheme, which corresponds to 16 equal divisions of each wedge and an angular resolution \ang{3.75}. In particular, the same or higher level of angular precision is achievable in NeuroHex with integer arithmetic alone. 

Polar domain operations in the Cartesian Coordinate System are bound to floating-point or fixed-point arithmetic. To ensure a fair comparison with NeuroHex, the Cartesian sine and cosine operations are implemented in fixed point with Bhāskara I's sine approximation. The number of fractional bits is selected to align with the angular resolution of Neurohex. The compute kernel applies a standard 2D rotation matrix to the input coordinates, after which the resulting values are rounded to the nearest integer. 

Rotation in NeuroHex follows a fundamentally different approach. The natural wedge-based partition enables the use of compact lookup tables, typically made of 6 or 7 entries to efficiently look up the sign configuration and positional ordering of the \textit{q}, \textit{r}, and \textit{s} cubic coordinates. Within each wedge, one and only one of the absolute values of \textit{q}, \textit{r}, and \textit{s} increases in consecutive units of step from 0 along the clock-wise direction of a hex ring, while one and only one remains constant along a given hex ring and increases radially in unit step from the origin across successive rings. We describe the former as the ring wedge spot numerator ($R_{ws}$) and the latter as the ring denominator. The positional relationship between the numerator and the denominator across the six wedges is defined as the \(\{ 1st\&2nd, \space 3rd\&1st, \space 2nd\&3rd, \space 1st\&2nd, \space 3rd\&1st, \space 2nd\&3rd \}\) elements of the cubic coordinates.

Rotating a point around an arbitrary center in the hexagonal grids is equivalent to translating that point by a specific distance along the hexagonal ring on which it resides. Consequently, the radial distance, or the ring denominator, of the initial and destination points remains unchanged during and after the rotation. For any rotation within a positive \ang{360} range, the operation can be decomposed into the \{number of whole wedges + the $R_{ws}$ remainder \}. In a 4-bit quantization scheme, for example, a rotation of \ang{150} corresponds to 50 quantization units moved along the hex ring that can be expressed as \(\ \lfloor \frac{50}{16} \rfloor = 3\) full wedges rotated with a residual displacement of \(50 - 3\cdot16=2\) quantization units along the ring. 

Subsequently, the process of NeuroHex rotation can be generalized into three steps. First, the ordered wedge index of the initial point is added to the number of whole wedges traversed during rotation to obtain the preliminary result wedge. Second, $R_{ws}$ of the initial point is added to the denormalized $R_{ws}$ of the rotational angle remainder, after which it is checked to determine whether the combined value exceeds the amount of a single wedge on the denormalized ring. The result wedge is incremented by 1 if an overflow occurs and reduced by 6 if the traversal extends the unit circle. The excess full wedge portion of the remainder, if present, is also removed. Finally, the new $R_{ws}$ is calculated within the result wedge. The values of \textit{q}, \textit{r}, and \textit{s} and their signs are assigned according to the location of the wedge.

\subsubsection{Orientation Predicate}
The orientation predicate refers to the orientation test in computational geometry, which determines whether a point lies on the clockwise or counterclockwise side of a directed line. In the Cartesian coordinate system, it is typically evaluated through the determinant calculated from a square matrix formed by the coordinates of the point of interest and two points from the line.

The test becomes somewhat more complicated in NeuroHex. The origin of the directed line is chosen first as a local reference point, and the coordinate values of the remaining two points are recalculated relative to the new origin. The second point defining the directed line can then be assigned deterministically to one of the six wedges by inspecting the signs of its coordinates, which is referred to as the line wedge. With the new canvas established, the position of the point of interest relative to the line may be categorized into three distinct scenarios:
\begin{enumerate}
    \item it lies within the two wedges immediately clockwise from the line wedge,
    \item it lies within the two wedges immediately counterclockwise from the line wedge, or
    \item it lies in either the same wedge as the line wedge or the opposite wedge.
\end{enumerate}
The first two cases, which together statistically account for $\frac{2}{3}$ of all possible situations, can be easily resolved by comparing the sign-based wedge index of the translated point of interest against a small lookup table. In the last $\frac{1}{3}$ of the situation when the point lies on the same or opposite wedge, the process requires comparisons of denormalized $R_{ws}$, which constitutes the critical path in the NeuroHex orientation predicate.

\subsubsection{Point in Sector Test}
The point in sector (PIS) test determines whether a point lies within a circular sector. A sector is defined by a center and two bounding infinite rays. To perform the test in the Cartesian coordinate system, the two boundary rays are converted into direction vectors with sine and cosine functions and checked against the sign of the 2D cross product formed between the direction vectors and the point vector relative to the sector center. 

One of the principal advantages of NeuroHex is that polar operations require substantially less arithmetic computation. For the PIS test, three examinations are needed to determine the result:
\begin{enumerate}
    \renewcommand{\labelenumi}{(\theenumi)}
    \item if the point lies in the same wedge as the left boundary ray, its $R_{ws}$ must be greater than or equal to that of the left boundary ray,
    \item if the point lies in the same wedge as the right boundary ray, its $R_{ws}$ must be smaller than or equal to that of the right boundary ray, and
    \item the wedge containing the point must be the same as, or sandwiched between, the ray wedges.
\end{enumerate}
The first two evaluations require denormalization of the angular $R_{ws}$ for comparison, whereas the final evaluation only requires proper handling of wrap-around cases. None of these steps involves trigonometry or fixed-point arithmetic.

\subsubsection{Point in Triangle Test}
As discussed in the shape library section, a triangle in NeuroHex can be constructed by intersecting two wedges so that one of their edges overlaps. Consequently, the PIT test becomes particularly straightforward, requiring only two instances of the PIS tests. The strategy may also be extended to more general polygons. On the other hand, in Cartesian coordinate systems, the PIT test is typically performed through three edge-function evaluations, one for each side of the triangle.

\subsection{Hardware Integration with Zynq-7000 FPGA}
The NeuroHex and Cartesian compute kernels are implemented with Vitis HLS design flow targeting the Xilinx Zynq-7000 FPGA platforms. Each kernel is embedded within a wrapper that stages input operands from AXI interface into local BRAM buffers, delivers the data to the kernels, stores the output arrays, and writes the results back to the host over AXI. Synthesis settings are chosen to minimize DSP usage in favor of fabric-only implementation. Pipelining is also deliberately disabled to maintain a streamlined and traceable implementation that aligns more closely with the extracted arithmetic operation counts of the geometric functions.

Once synthesized, the compute kernels are packaged in Vitis and exported as IP cores for Vivado. These IPs are instantiated within a Vivado block diagram and connected to the Zynq processing system through AXI interconnects, after which the complete system is implemented and a bitstream is generated. The hardware configuration is then exported as an .xsa file that is used to create a platform project in Vitis and build the application project. The host executable uses DMA to access the assigned AXI address space, programs the FPGA, transfers the kernel input, and invokes the kernel wrapper. During deployment, the target board is booted from an SD card containing a Petalinux system image, a bitstream file, and the executable. At runtime, the FPGA is first programmed with the generated bitstream before executing the host application to launch the kernels. 

\subsection{Discussion on Hardware Implementation Results}
Table~\ref{tab:synthesis_report} presents the DSP-free synthesis results generated in Vitis HLS. Table~\ref{tab:implementation_report} reports the post-implementation resource utilization and power estimate from Vivado. The breakdown of the arithmetic operations invoked by each kernel is provided in Table~\ref{tab:operation_count}. In general, for geometric shapes that can be localized to wedge/ring coordinates and evaluated through low-width ratio-based comparisons, NeuroHex is superior to Euclidean arithmetic by a huge margin, yielding approximately 2x to 12x resource reduction than their Cartesian counterparts for distance, polar angle, rotation, and PIS testing. These characteristics make NeuroHex well-suited for reference frames, as rotational and polar operations are prevalent, and offer substantial resource savings for hardware implementation. For tasks that can be reduced naturally to one or more signed 2D determinants, Cartesian arithmetic is more advantageous. However, even for operations such as the orientation predicate, NeuroHex is more efficient in $\frac{2}{3}$ of the cases, since the early clockwise and counter-clockwise classifications can often be resolved after only a few comparisons. NeuroHex can also be aggressively scaled down to a lower bitwidth as a result of wedge-based arithmetic, but this implementation keeps data types to int8, int16, and int32 for the sake of universality.

It is also important to note that the additional coordinate in the NeuroHex coordinates increases memory traffic and demands more interconnects and storage, which is reflected in the comparatively heavier fabric utilization of operations that require higher counts of coordinates as input. One practical mitigation is to transfer only two hex coordinates as input and perform two additional subtraction/additions to reconstruct the third coordinates when needed. More broadly speaking, if storage is strategically planned to contain only the signed ring wedge spot numerator and ring denominator, the compute kernels will become even more simplified and memory traffic will reduce significantly.

\begin{table}[t]
  \centering
  \begin{tabular}{lrr}
    \toprule
    Kernel & FF & LUT \\
    \midrule
    NeuroHex Distance & 50 & 240 \\
    Cartesian Distance & 104 & 885 \\
    \cmidrule(lr){1-3}
    NeuroHex Polar Angle & 53 & 352 \\
    Cartesian Polar Angle & 821 & 1419 \\
    \cmidrule(lr){1-3}
    NeuroHex Rotate & 142 & 660 \\
    Cartesian Rotate & 1057 & 4797 \\
    \cmidrule(lr){1-3}
    NeuroHex Orientation Predicate & 154 & 775 \\
    Cartesian Orientation Predicate & 53 & 392 \\
    \cmidrule(lr){1-3}
    NeuroHex PIS Test & 53 & 526 \\
    Cartesian PIS Test & 1729 & 6226 \\
    \cmidrule(lr){1-3}
    NeuroHex PIT Test & 184 & 1019 \\
    Cartesian PIT Test & 35 & 933 \\
    \bottomrule
  \end{tabular}
  \caption{Post-synthesis Resource Utilization of NeuroHex and Cartesian Compute Kernels from Vitis suggests that the NeuroHex implementation is most competent for primitives dealing with angle displacement, averaging 4 to 10 times less FPGA fabric usage. On the other hand, Cartesian kernels are more cost-efficient for operations that exploit edge functions, with a slight lead on PIT test and 2x resource efficiency for orientation predicate. }
  \label{tab:synthesis_report}
\end{table}

\begin{table*}[t]
  \centering
  \scriptsize
  \setlength{\tabcolsep}{5pt}
  \begin{tabular}{lrrrrrrrr}
    \toprule
    Name & Slice LUTs & Slice Registers & F7 Muxes & F8 Muxes & Slice & LUT as Logic & DSP & Power (W)\\
    \midrule
    NeuroHex Distance & 28 & 50 & 0 & 0 & 20 & 28 & 0 & <0.001\\
    Cartesian Distance & 191 & 91 & 0 & 0 & 96 & 191 & 0 & 0.004\\ 
    \cmidrule(lr){1-9}
    NeuroHex Polar Angle & 54 & 53 & 0 & 0 & 0 & 0 & 0 & 0.001\\
    Cartesian Polar Angle & 351 & 475 & 0 & 0 & 0 & 0 & 0 & 0.003\\
    \cmidrule(lr){1-9}
    NeuroHex Rotation & 143 & 84 & 0 & 0 & 0 & 0 & 0 & 0.001\\
    Cartesian Rotation & 830 & 465 & 0 & 0 & 0 & 0 & 0 & 0.001\\
    \cmidrule(lr){1-9}
    NeuroHex Ori Pred & 313 & 148 & 0 & 0 & 113 & 313 & 0 & 0.003\\
    Cartesian Ori Pred & 80 & 53 & 0 & 0 & 61 & 80 & 0 & 0.001\\
    \cmidrule(lr){1-9}
    NeuroHex PIS Test & 35 & 45 & 0 & 0 & 22 & 35 & 2 & 0.001\\
    Cartesian PIS Test & 1270 & 778 & 0 & 0 & 425 & 1266 & 0 & 0.001\\
    \cmidrule(lr){1-9}
    NeuroHex PIT Test & 141 & 80 & 0 & 0 & 99 & 141 & 4 & <0.001\\
    Cartesian PIT Test & 177 & 35 & 0 & 0 & 142 & 177 & 0 & 0.003\\
    \bottomrule
  \end{tabular}
  \caption{Post-implementation Resource Utilization and Power Report from Vivado displays similar trends to those shown in Table 2, taking into account the added usage of DSP. The power report indicates that NeuroHex outperforms Cartesian with a steady 3x-4x less power consumption in most cases but falls behind on orientation predicate. }
  \label{tab:implementation_report}
\end{table*}

\begin{table*}[t]
  \centering
  \scriptsize
  \setlength{\tabcolsep}{2.8pt}
  \renewcommand{\arraystretch}{1.05}
  \resizebox{0.7\textwidth}{!}{%
  \begin{tabular}{ll|cc|cc|cc|cc|cc|cc}
    \toprule
    \multirow{2}{*}{Kernel} & \multirow{2}{*}{Type} &
    \multicolumn{2}{c|}{Add} &
    \multicolumn{2}{c|}{Sub} &
    \multicolumn{2}{c|}{Mul} &
    \multicolumn{2}{c|}{Div} &
    \multicolumn{2}{c|}{Cmp} &
    \multicolumn{2}{c}{Sqrt} \\
    & &
    NH & C & NH & C & NH & C & NH & C & NH & C & NH & C \\
    \midrule

    \multirow{4}{*}{\texttt{op\_distance}}
      & combined      & \heat{0} & \heat{2} & \heat{6} & \heat{2} & \heat{0} & \heat{4} & \heat{0} & \heat{0} & \heat{8} & \heat{0} & \heat{0} & \heat{1} \\
      & 16-bit        & \heat{0} & \heat{0} & \heat{6} & \heat{2} & \heat{0} & \heat{2} & \heat{0} & \heat{0} & \heat{8} & \heat{0} & \heat{0} & \heat{1} \\
      & 32-bit        & \heat{0} & \heat{1} & \heat{0} & \heat{0} & \heat{0} & \heat{2} & \heat{0} & \heat{0} & \heat{0} & \heat{0} & \heat{0} & \heat{0} \\
      & 10-bit fixed  & \heat{0} & \heat{1} & \heat{0} & \heat{0} & \heat{0} & \heat{0} & \heat{0} & \heat{0} & \heat{0} & \heat{0} & \heat{0} & \heat{0} \\
    \midrule

    \multirow{3}{*}{\texttt{op\_polar\_angle}}
      & combined   & \heat{1} & \heat{6} & \heat{2} & \heat{6} & \heat{1} & \heat{4} & \heat{0} & \heat{4} & \heat{7} & \heat{14} & \heat{0} & \heat{0} \\
      & 8-bit      & \heat{1} & \heat{0} & \heat{2} & \heat{0} & \heat{1} & \heat{0} & \heat{0} & \heat{0} & \heat{7} & \heat{0} & \heat{0} & \heat{0} \\
      & 16-bit     & \heat{0} & \heat{6} & \heat{0} & \heat{6} & \heat{0} & \heat{4} & \heat{0} & \heat{4} & \heat{0} & \heat{14} & \heat{0} & \heat{0} \\
    \midrule

    \multirow{5}{*}{\texttt{op\_rotate}}
      & combined     & \heat{3} & \heat{5} & \heat{10} & \heat{14} & \heat{1} & \heat{6} & \heat{0} & \heat{2} & \heat{10} & \heat{12} & \heat{0} & \heat{0} \\
      & 8-bit        & \heat{1} & \heat{0} & \heat{10} & \heat{0} & \heat{1} & \heat{0} & \heat{0} & \heat{0} & \heat{10} & \heat{0} & \heat{0} & \heat{0} \\
      & 16-bit       & \heat{2} & \heat{2} & \heat{0} & \heat{9} & \heat{0} & \heat{2} & \heat{0} & \heat{0} & \heat{0} & \heat{10} & \heat{0} & \heat{0} \\
      & 16-bit fixed & \heat{0} & \heat{0} & \heat{0} & \heat{2} & \heat{0} & \heat{0} & \heat{0} & \heat{2} & \heat{0} & \heat{0} & \heat{0} & \heat{0} \\
      & 28-bit fixed & \heat{0} & \heat{3} & \heat{0} & \heat{3} & \heat{0} & \heat{4} & \heat{0} & \heat{0} & \heat{0} & \heat{2} & \heat{0} & \heat{0} \\
    \midrule

    \multirow{4}{*}{\texttt{op\_ori\_pred}}
      & combined   & \heat{0} & \heat{0} & \heat{13} & \heat{5} & \heat{2} & \heat{4} & \heat{0} & \heat{0} & \heat{24} & \heat{2} & \heat{0} & \heat{0} \\
      & 8-bit      & \heat{0} & \heat{0} & \heat{0} & \heat{0} & \heat{2} & \heat{0} & \heat{0} & \heat{0} & \heat{4} & \heat{0} & \heat{0} & \heat{0} \\
      & 16-bit     & \heat{0} & \heat{0} & \heat{13} & \heat{4} & \heat{0} & \heat{2} & \heat{0} & \heat{0} & \heat{20} & \heat{0} & \heat{0} & \heat{0} \\
      & 32-bit     & \heat{0} & \heat{0} & \heat{0} & \heat{1} & \heat{0} & \heat{2} & \heat{0} & \heat{0} & \heat{0} & \heat{2} & \heat{0} & \heat{0} \\
    \midrule

    \multirow{6}{*}{\texttt{op\_pis}}
      & combined     & \heat{0} & \heat{5} & \heat{6} & \heat{26} & \heat{2} & \heat{13} & \heat{0} & \heat{4} & \heat{13} & \heat{28} & \heat{0} & \heat{0} \\
      & 8-bit        & \heat{0} & \heat{0} & \heat{0} & \heat{0} & \heat{2} & \heat{1} & \heat{0} & \heat{0} & \heat{2} & \heat{0} & \heat{0} & \heat{0} \\
      & 16-bit       & \heat{0} & \heat{4} & \heat{6} & \heat{20} & \heat{0} & \heat{6} & \heat{0} & \heat{0} & \heat{11} & \heat{20} & \heat{0} & \heat{0} \\
      & 32-bit int   & \heat{0} & \heat{1} & \heat{0} & \heat{0} & \heat{0} & \heat{2} & \heat{0} & \heat{0} & \heat{0} & \heat{2} & \heat{0} & \heat{0} \\
      & 16-bit fixed & \heat{0} & \heat{0} & \heat{0} & \heat{4} & \heat{0} & \heat{0} & \heat{0} & \heat{4} & \heat{0} & \heat{0} & \heat{0} & \heat{0} \\
      & 28-bit fixed & \heat{0} & \heat{0} & \heat{0} & \heat{2} & \heat{0} & \heat{4} & \heat{0} & \heat{0} & \heat{0} & \heat{6} & \heat{0} & \heat{0} \\
    \midrule

    \multirow{4}{*}{\texttt{op\_pit}}
      & combined   & \heat{0} & \heat{0} & \heat{12} & \heat{15} & \heat{4} & \heat{6} & \heat{0} & \heat{0} & \heat{26} & \heat{3} & \heat{0} & \heat{0} \\
      & 8-bit      & \heat{0} & \heat{0} & \heat{0} & \heat{0} & \heat{4} & \heat{0} & \heat{0} & \heat{0} & \heat{4} & \heat{0} & \heat{0} & \heat{0} \\
      & 16-bit     & \heat{0} & \heat{0} & \heat{12} & \heat{12} & \heat{0} & \heat{6} & \heat{0} & \heat{0} & \heat{22} & \heat{0} & \heat{0} & \heat{0} \\
      & 32-bit int & \heat{0} & \heat{0} & \heat{0} & \heat{3} & \heat{0} & \heat{0} & \heat{0} & \heat{0} & \heat{0} & \heat{3} & \heat{0} & \heat{0} \\
    \bottomrule
  \end{tabular}%
  }
  \caption{Arithmetic Operation Counts for NeuroHex (NH) and Cartesian (C) Compute Kernels With 4-bit Quantization. The algorithmic structure of geometric operations is consistent with the observed utilization of resources. NeuroHex excels at converting geometric problems into integer selection, wedge localization, and bounded comparison.}
  \label{tab:operation_count}

\end{table*}

\section{Experimentation with Real-World Spatial Data}

\begin{figure*}
            \centering
            \includegraphics[width=0.78\linewidth]{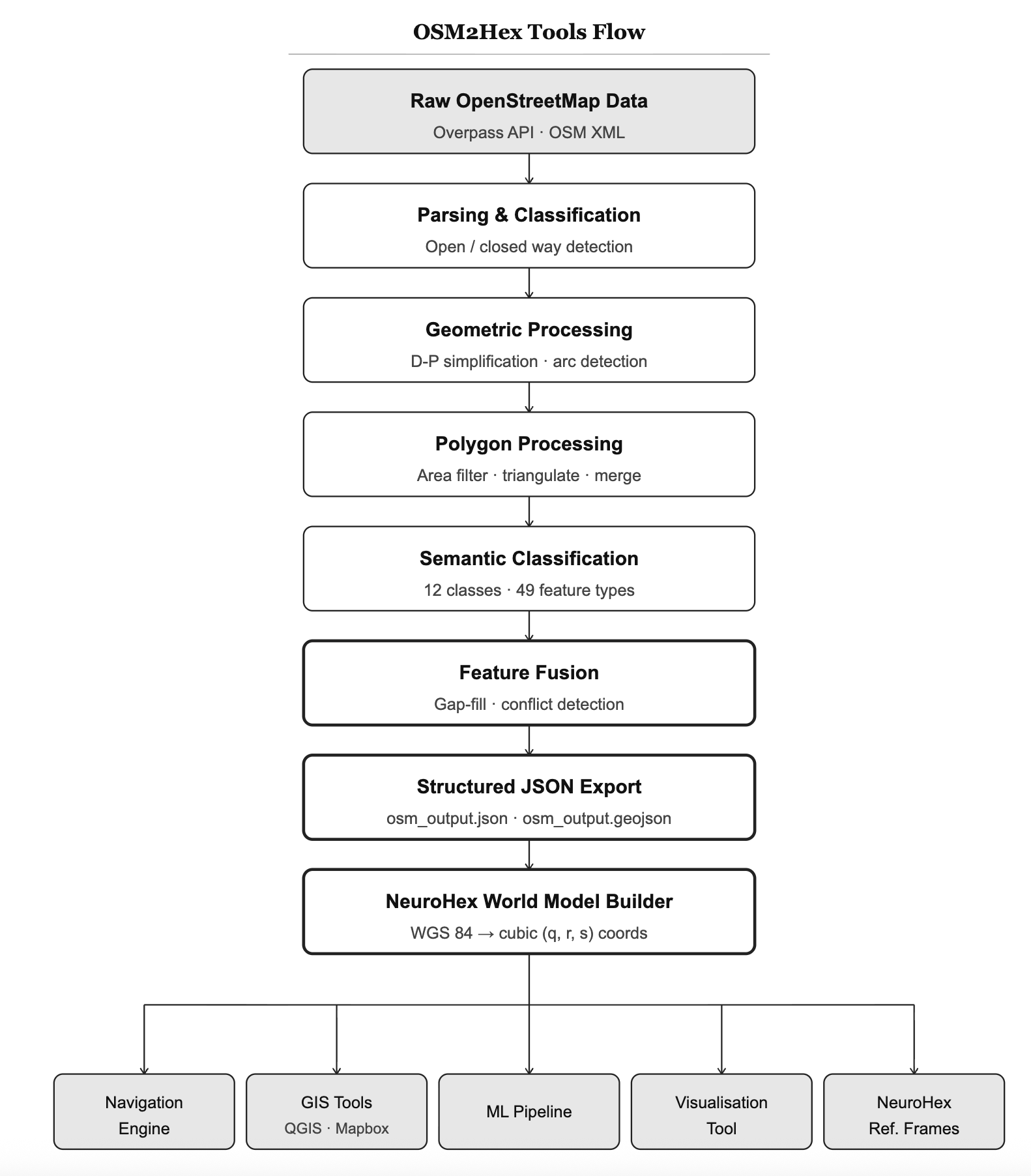}
            \caption{Complete end-to-end data flow of the OSM2Hex system. Raw
    OpenStreetMap data is acquired via the Overpass API, parsed, classified, geometrically
    simplified, triangulated, and merged into NeuroHex-compatible shape primitives. At the feature level primitives are exported as a
    structured JSON file consumed by the NeuroHex world model builder and downstream
    applications.}
            \label{fig:full_pipeline}
\end{figure*}

\begin{figure*}
        \centering
        \includegraphics[width=\linewidth,height=0.9\textheight,keepaspectratio]{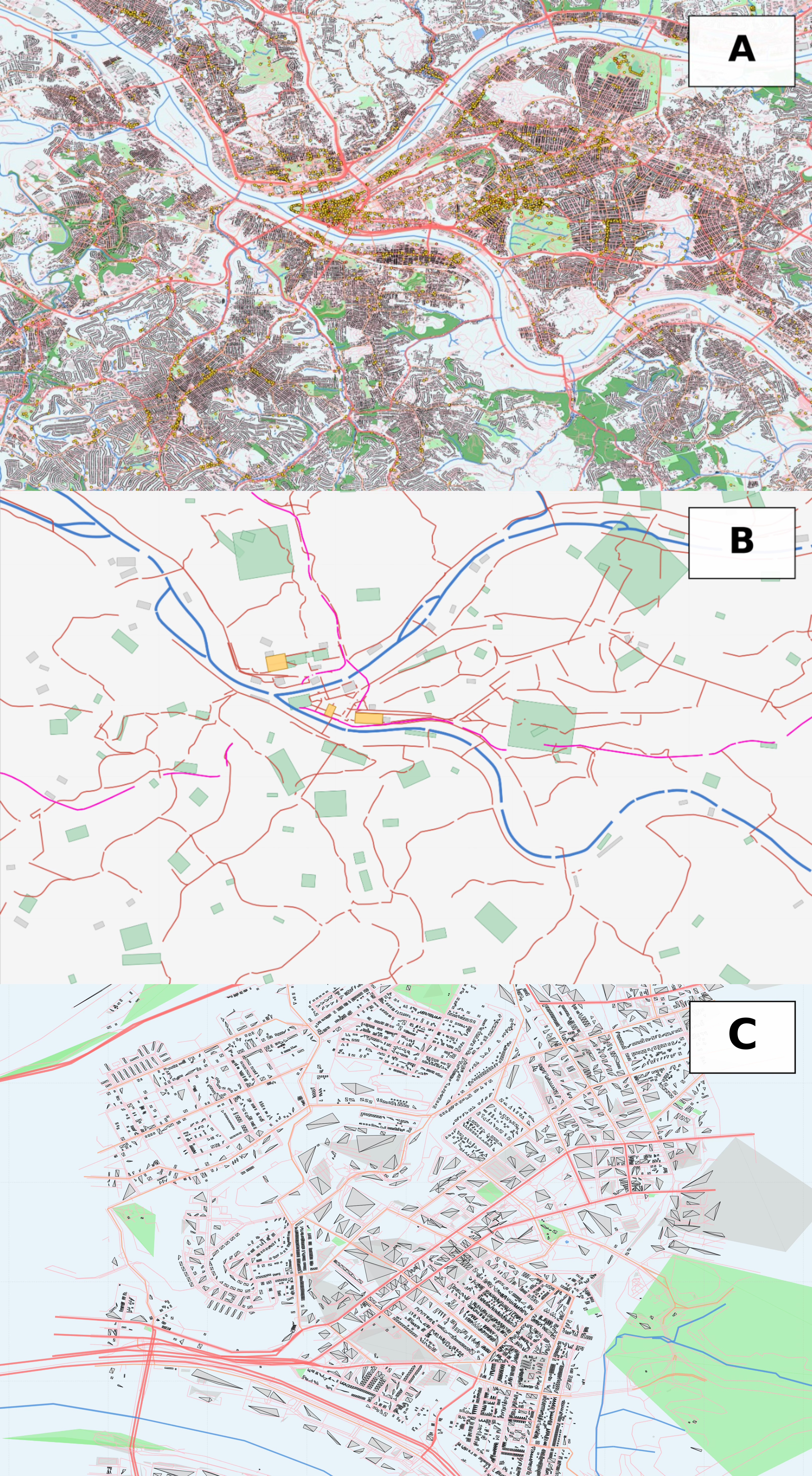}
        \caption{\centering \textit{OSM2Hex} conversion tool: (A) Raw OpenStreetMap data for the Pittsburgh metro area. (B) Pittsburgh metro area filtered and replaced with NeuroHex primitives. (C) Zoomed in Oakland area with higher resolution NeuroHex primitives.}
    \label{fig:Pittsbrugh_maps}
\end{figure*}

For NeuroHex to support spatial data in the real world at scale, we develop a spatial data processing tool that transforms OpenStreetMap (OSM) \cite{OSM} data into our NeuroHex framework, analogous to the Sensory Processing Pipeline shown in Figure~\ref{fig:Top-Level Model}. 

We use this OSM-to-NeuroHex conversion tool (\textit{OSM2Hex}) to better understand the efficiency of NeuroHex in encoding real-world information. Looking at real spatial data also helps us to scope the needed sizing of working memory and reference frames in long-term memory going forward. 
Figure~\ref{fig:full_pipeline} shows the complete end-to-end processing pipeline of
the OSM2Hex conversion tools flow.

This section evaluates NeuroHex using OSM2Hex and geographic data for the Pittsburgh metropolitan area and a higher-resolution analysis of the Oakland area (containing Carnegie Mellon University and University Of Pittsburgh campuses). These experiments allow us to empirically determine parameters for reference frame size, object limits, geometric simplification thresholds, and multi-scale resolution behavior.

\subsection{From Raw OSM to NeuroHex Spatial Abstraction}

We begin with a full extraction of OpenStreetMap (OSM) data covering the Pittsburgh metropolitan area (Figure \ref{fig:Pittsbrugh_maps}.A). To support scalable real-world spatial abstraction, we use OSM2Hex conversion tool to transform raw OSM geometries into a compact, navigation-relevant NeuroHex world model. Given a user-defined geographic bounding box, OSM2Hex parses OSM XML data into primitive geometric candidates (polylines, polygons, and nodes) and groups them by semantic class using OSM tags (e.g., rivers, transportation corridors, buildings, parks, and water features).

Each feature class is processed using a parameterized geometric simplification pipeline. Linear features are simplified using Douglas--Peucker reduction~\cite{douglas1973algorithms} to preserve global topology while reducing vertex density, followed by Chaikin smoothing~\cite{chaikin1974algorithm} to improve perceptual continuity. Polygonal features undergo area-based filtering, vertex capping, and boundary smoothing to remove low-impact structures while retaining salient landmarks. The features are further filtered according to relevance.
The remaining geometries are replaced with primitive mosaics compatible with NeuroHex composed of straight-line and arc segments and a small palette of simple polygonal shapes. The aggressiveness of filtering and primitive substitution is controlled by a resolution policy
enabling adaptive trade-offs between abstraction and detail.

The raw OSM extraction for the Pittsburgh metropolitan area contains approximately 27{,}000 road polylines, 18{,}500 building polygons, and 2{,}300 park and water features, totaling roughly 45{,}000 geometries. After normalization, feature grouping, and geometric preprocessing using Douglas--Peucker simplification~\cite{douglas1973algorithms} and Chaikin filtering~\cite{chaikin1974algorithm}, average polyline vertex counts are reduced by approximately 82\%, and polygon vertex counts are reduced from an average of 95 to 21. Features below an area threshold of $2.5 \times 10^{-6}\,\text{deg}^2$ (e.g., small sheds, garages, and interior courtyards) are completely removed. Following preprocessing, the dataset is reduced to approximately 11{,}200 meaningful features, representing a 75\% reduction relative to the original OSM geometry count.

\subsection{Relevance-Based Feature Prioritization}

To further compress the data set while maintaining a meaningful navigation structure, we apply a hierarchical three-stage filtering process. All line-based features are converted into NeuroHex-compatible straight-line and arc primitives:

\begin{enumerate}
    \item \textit{Identity Features:} Primary rivers (Allegheny, Monongahela, Ohio) are merged from 85 polyline fragments into 14 stylized splines, representing an 84\% reduction.
    \item \textit{Structural Features:} Major highways and arterial roads are filtered and simplified from 9,400 polyline segments to 720 primitives (92\% reduction).
    \item \textit{Contextual Features:} Building footprints and landmarks are reduced from 18,500 polygons to 1,200 composite structures (93\% reduction).
\end{enumerate}
This process intentionally preserves the features humans rely on for orientation---rivers, major corridors, and landmark clusters---while removing redundant or low-impact geometries. After filtering, the prioritized navigational structure consists of 1,934 key objects (14 identity features, 720 structural features, and 1,200 contextual features), a reduction of roughly 96\% relative to the original 45,000 raw OSM geometries.

\subsection{Shape Abstraction Using NeuroHex Primitive Mosaics}

Each object can be represented as a mosaic of NeuroHex simple shapes selected from a small palette: rectangles, circles, and triangles. Each polygonal object is replaced by 2 to 8 primitives, depending on the area and perceptual relevance.
On average: 1) Parks use 12 to 20 primitives; 2) Building clusters use 3 to 5 rectangles or triangles; and 3) Landmarks use 4 to 9 blended primitives.
Across the filtered data set of 1,934 objects, mosaic abstraction resulted in approximately 12,000 total NeuroHex simple shapes, corresponding to another 70\% reduction in geometric complexity relative to raw polygon and vertex counts. The resultant  abstraction is shown in Figure \ref{fig:Pittsbrugh_maps}.B.

The 1,934 objects on the metro scale exceed the number of objects humans can maintain simultaneously. Cognitive studies suggest that people manage mental maps containing dozens to hundreds of landmarks~\cite{burgess2002human}. This motivates dividing the metro region into hierarchical reference frames. This is something people do naturally. The Pittsburgh metro area consists of multiple townships, and the City of Pittsburgh itself contains more than 90 neighborhoods. Strong evidence that people logically divide their world down to units they can hold in reference frames and their working memory.

\subsection{High-Resolution Example: Pittsburgh's Oakland Area}

NeuroHex naturally supports zooming adaptive spatial resolutions. To evaluate fine-grained behavior in a dense urban environment, we apply the OSM2Hex conversion tool to the Oakland area of Pittsburgh (Figure \ref{fig:Pittsbrugh_maps}.C). This area contains two  university campuses (Carnegie Mellon University and the University of Pittsburgh), professional, retail, residential, and green-spaces representative of high-density urban subregions. Fine-grained navigable information for a region like this can provide insight to explore how spatial data can be organized into reference frames and working memory.

We use the same algorithms to reduce the complexity of the OSM data for this subregion, but adapt the relevance to preserve more information as we zoom in. The metro-scale policy retains only identity and structural features, yielding a sparse corridor-level abstraction, whereas the zoom-scale policy preserves dense contextual features such as individual buildings and paths, resulting in a several-fold increase in feature and primitive density over the same geographic area. The OSM2Hex conversion tool ultimately converts all spatial data down to simple NeuroHex shapes. For this subregion, the polyline data (roads, paths, rivers) can be represented with 5,599 NeuroHex primitives (line segments, and arcs). The polygon data (parks, buildings, administrative boundary) consist of 4,612 polygons. The geometry of parks on average takes approximately 2.7 simple NeuroHex shapes. The geometry of many buildings can be  captured with a single quadrilateral NeuroHex shape. Some complex buildings may require tens of simple NeuroHex shapes. With our current fidelity, we average 2 simple NeuroHex shapes per building. Overall, we used 9,237 NeuroHex shapes to represent the polygon data. Overall, this large complex urban area shown in Figure \ref{fig:Pittsbrugh_maps}.C can be encoded in less than 15,000 simple NeuroHex shapes.

\subsection{Summary of Applying NeuroHex on Real World Data Sets}

For Pittsburgh city and Oakland neighborhood, OSM2Hex conversion tool successfully reduces real-world spatial data through successive stages of geometric simplification, relevance filtering, primitive substitution down to NeuroHex simple shapes. For the metro-scale, we see over two orders of magnitude in reduction of geometric information down to under 12,000 simple NeuroHex shapes, while still providing information sufficient for general orientation and even navigation between neighborhoods. For the zoomed-in Oakland neighborhood, we preserve all the roads and even pedestrian paths along with nearly all buildings for this subregion, resulting in a detailed navigable map encoded in less than 15,000 simple NeuroHex shapes.

These empirical results provide quantitative guidance for determining spatial coordinate bounds, reference-frame object limits, simplification tolerances, and abstraction granularity. Together, they demonstrate that large-scale environments can be efficiently encoded using compact and human-interpretable visual primitives suitable for hierarchical spatial memory map and intelligent adaptive navigation within it. The OSM2Hex conversion framework takes advantage of publicly available OSM data sets for the creation of spatial maps of arbitrary geographic areas based on the NeuroHex coordinate system. Using OSM2Hex we can quickly generate the initial world map of a chosen physical area at varying degrees of resolutions. Such world maps can be dynamically updated to adapt to potentially changing environments.

\section{Summary}

We present a novel hexagonal coordinate system that offers a highly efficient substrate
for building dynamic world models to enable adaptive spatial reasoning and continuous online-adaptive learning (COAL).
Inspired by grid cells in the human brain, \textit{NeuroHex} employs a hexagonal isometric coordinate formulation that provides rotational symmetry and constant-cost translation, rotation, and distance computation. The NeuroHex framework incorporates ring indexing, quantized angular encoding, and a hierarchical library of foundational, simple, and complex geometric shape primitives. 
We also developed \textit{OSM2Hex}, a tool for converting OpenStreetMap (OSM) data sets 
into NeuroHex world models. We show example results using the city of Pittsburgh.
The results demonstrate that NeuroHex offers a highly computationally efficient platform for online world-model construction and adaptive spatial reasoning. NeuroHex platforms can enable edge-native, energy-efficient AI capabilities for autonomous and online-adaptive perception, cognition, and navigation applications.

\bibliographystyle{IEEEtran}
\bibliography{refs}

\end{document}